\documentclass[letterpaper,10pt,conference]{ieeeconf}

\IEEEoverridecommandlockouts
\usepackage[utf8]{inputenc}
\usepackage{amsmath,amssymb,amsfonts}
\usepackage{mathtools}
\usepackage{bm}

\usepackage{graphicx}
\usepackage{cite}
\usepackage{xcolor}
\usepackage{booktabs}
\usepackage{multirow}
\usepackage{makecell}
\usepackage{array}
\usepackage{url}
\usepackage[normalem]{ulem}
\usepackage{algorithmic}
\usepackage{textcomp}

\usepackage{tikz}
\usepackage{pgf}

\usetikzlibrary{
    arrows.meta,
    positioning,
    calc,
    fit,
    shapes.geometric,
    shapes.multipart,
    shapes.symbols,
    backgrounds
}

\def\BibTeX{{\rm B\kern-.05em{\sc i\kern-.025em b}\kern-.08em
    T\kern-.1667em\lower.7ex\hbox{E}\kern-.125emX}}

\usepackage[caption=false,font=footnotesize]{subfig}

\usepackage[linesnumbered,ruled,vlined]{algorithm2e}

\usepackage[hidelinks]{hyperref}

\usepackage{accents}
\newlength{\dhatheight}

\begin{document}

\title{\LARGE \bf Splat-CBF: Safe Next-Best-View Control in 3D Gaussian-Splat Maps}


\author{Amirhossein Mollaei Khass, Athanasios Cosse, Nader Motee
\thanks{
This work was supported in part by the ONR under grants numbers  N00014-23-1-2779 and N00014-26-1-2246. \newline
A.M. Khass, A. Cosse, and N. Motee are with the Department of Mechanical Engineering and Mechanics, Lehigh University, Bethlehem, PA, 18015, USA. {\tt\small \{ammb23,asc425,motee\}@lehigh.edu}.\endgraf
}
}

\maketitle

\begin{abstract}

Where to look and how to move? A robot navigating an unmapped environment must do both at once, and the two goals pull against each other. The regions most worth observing are the ones the map knows least about, and those are exactly where the robot cannot trust its collision margins. We resolve this tension by introducing Splat-CBF, an active perception control barrier function that steers the camera toward the next best view while collision avoidance is enforced as a hard constraint. Safety is enforced by a risk-aware control barrier function that turns the Average Value-at-Risk of the Gaussian field into a single smooth hard constraint. Perception is enforced by a second barrier that rewards camera orientations with high expected Fisher information gain near the robot's planned path. The two meet in a quadratic program where safety is hard and perception is soft, with a slack penalty that adapts to how often perception has already been relaxed and how close the robot is to an uncertain region. We verify the method in indoor simulations, a Isaac Kinova manipulator and in experiments on an Ackermann-drive robot. Our results assert that robot navigates faster, gathers more information, and runs faster online than safety-only and perception-only baselines, giving up informative motion only when safety requires it.

\end{abstract}
%



\section{Introduction}
\label{sec:introduction}

Autonomous robots operating in unknown or partially observed environments must navigate
safely toward a goal while acquiring observations that improve future decisions. 3D
Gaussian Splatting (3DGS) provides a dense, differentiable, and efficiently renderable
scene representation for this setting~\cite{kerbl20233d}, in which safe autonomy requires the robot to select actions that are simultaneously safe and informative~\cite{chen2025splat}.

Control barrier functions (CBFs) enforce safety constraints on a control
policy~\cite{cbf_aron2019}, and recent 3DGS navigation methods add geometric and CBF-based
safety filters~\cite{chen2025control,dario2026fastbridge,hong2026polymerge}. Dense
Gaussian maps, however, make per-obstacle constraints expensive, and safety alone does not
reduce the uncertainty that governs future motion, which leaves the robot overly
conservative~\cite{khass2026conflict}. Next-best-view (NBV) and active perception methods
take the other side. They select informative views or trajectories to improve
reconstruction~\cite{jiang2024fisherrf}, but do not enforce safety during execution~\cite{khass2025active}, and they return informative trajectories rather than
safe control actions~\cite{tao2025rt,jiang2024ag}.

The two sides meet in unknown environments. The uncertain regions that dominate the
information objective are the same regions whose safety margins are least reliable. Thus,
the information objective pulls the robot toward poorly modeled regions, while the safety
constraint pushes it away from them. We resolve this tension by introducing Splat-CBF, an active perception CBF that selects
the next best view and is coupled with a hard collision avoidance constraint in a single
QP. Collision avoidance with Gaussian
splat obstacles is a hard constraint. It is coupled with a soft perception barrier that
drives the robot toward informative observations inside a trajectory-relevant mask around
its anticipated motion, rather than over the full map. When the two conflict, the
resulting CBF-QP relaxes the perception objective through an adaptive slack weight.

Our main contributions are: (i) a composite safety CBF that aggregates the collision risk
of nearby Gaussian primitives through Average Value-at-Risk; (ii) Splat-CBF, an active
perception CBF that promotes the expected Fisher information gain of a view and is coupled
with the safety CBF in a single QP, where collision avoidance is hard and perception is
soft; (iii) an adaptive slack weight that relaxes perception online when information gain
conflicts with collision avoidance; and (iv) a validation in indoor simulation and Isaac Kinova manipulator and in
hardware experiments on an Ackermann-drive, with online 3DGS reconstruction.






\section{Related work.}
\label{sec:related_work}

Recent 3DGS active perception methods use Fisher information and related uncertainty
measures to select views and trajectories. FisherRF~\cite{jiang2024fisherrf} formulates
active view selection through Fisher information, AG-SLAM~\cite{jiang2024ag} extends
information-aware planning to active 3DGS SLAM, and RT-GuIDE~\cite{tao2025rt} develops
efficient information-driven exploration, while risk-aware NBV methods add collision risk
to the EIG objective~\cite{khass2025active}. These approaches improve reconstruction and
informative planning, but they do not embed perception in safety-critical closed-loop control. On the safety side, Splat-Nav~\cite{chen2025splat} uses Gaussian geometry for
collision-aware planning, SAFER-Splat~\cite{chen2025control} imposes CBF constraints on
Gaussian obstacles, and recent methods improve efficiency through dynamics-aware filtering
or Gaussian-map compression~\cite{dario2026fastbridge,hong2026polymerge}. These methods
keep the robot safe, but they treat perception as a byproduct of motion rather than as an
objective. The closest method is \cite{khass2026conflict}, which aligns the camera with
uncertain regions through separate spatial and angular barriers with fixed weights.
Splat-CBF instead enforces the expected Fisher information gain of the view itself as a
single active perception barrier, and sets the weight of its relaxation online from the
conflict history and the proximity to unmapped regions. Simulations and hardware
experiments in Section~\ref{sec:experiments} show faster goal reaching, higher information
gain, and lower computation time than \cite{khass2026conflict}.



\section{Problem Formulation}
\label{sec:problem_formulation}

\begin{figure*}[t]
\centering
\includegraphics[width=\textwidth,trim={1.5cm 3.7cm 1.2cm 5.6cm},clip]{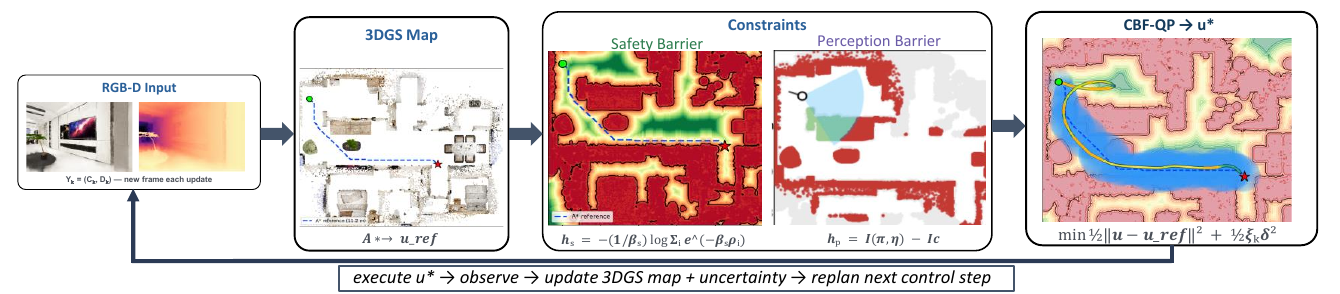}
\caption{System overview. Each RGB-D observation updates the 3DGS map and
nominal plan, while a unified CBF-QP combines a hard safety barrier with a
soft trajectory-aware perception barrier to generate safe, informative
control actions in closed loop.}

\label{fig:system_overview}
\vspace{-0.4cm}
\end{figure*}
We consider a robot that navigates toward a goal or completes a task in an environment with limited prior knowledge and an initially incomplete scene representation. At each interval update, the robot receives an onboard RGB-D observation and incrementally updates its online map. The objective is not merely to maximize global reconstruction quality, but to acquire observations that improve the geometric information relevant to future motion planning, collision avoidance, and task execution. This creates a safety--perception tradeoff; informative observations often require motion or camera reorientation toward occluded, unexplored, or uncertain regions, which may simultaneously have small clearance or uncertain collision risk. We therefore seek online control actions that maintain safety, make task progress, and improve the map in regions relevant to the robot's future motion.

We model the robot as a control-affine system
\begin{equation}
    \dot{x}=f(x)+g(x)u,
    \label{eq:dynamics}
\end{equation}
where $x\in\mathcal{X}\subset\mathbb{R}^{n}$ is the robot state and
$u\in\mathcal{U}\subset\mathbb{R}^{m}$ is the control input. Let
$\pi(x)\in\mathbb{R}^{d}$ and $\eta(x)\in\mathbb{S}^{d-1}$ denote the robot position and camera viewing direction, respectively.
The camera pose is denoted by $T= (\pi,\eta)$ and with map parameters $w_k$, the differentiable rendering model
\begin{equation}
    \widehat Y_k
    =
    \mathcal{F}\big(T,w_k\big),
    \qquad
    \widehat Y_k=(\widehat C_k,\widehat D_k),
    \label{eq:rendering_model}
\end{equation}
predicts an RGB-D image via alpha compositing of the projected Gaussian primitives~\cite{kerbl20233d}. Given an observed RGB-D image $Y_k=(C_k,D_k)$, the local map is updated online by
minimizing over image pixels $l$ 
\begin{equation}
\small
\begin{aligned}
    \mathcal L(w_k)
    =
    \sum_l
    \left[
        \mathcal L_1(C_{k,l},\widehat C_{k,l})
        +
        \psi\,\mathcal L_1(D_{k,l},\widehat D_{k,l})
    \right],
    \label{eq:3dgs_loss}
\end{aligned}
\end{equation}
for some $\psi\in[0,1)$. The environment is represented by an online 3D Gaussian Splatting map (3DGS)~\cite{kerbl20233d} $\mathcal{G}_k=\{g_i^k\}_{i=1}^{M_k},$
where each Gaussian primitive is written as
\begin{equation}
    g_i=(\mu_i,R_i,s_i,o_i,c_i), \qquad
    \Sigma_i=R_i\operatorname{diag}(s_i^2)R_i^\top .
    \label{eq:gaussian_map}
\end{equation}
Here $\mu_i\in\mathbb{R}^3$ is the mean position, $R_i\in SO(3)$ is the orientation, $s_i\in\mathbb R_{>0}^3$ is the scale, $o_i$ is opacity, $c_i$ denotes appearance.
Safety is specified through the safe set
\begin{equation}
    \mathcal{C}_s
    =
    \{x\in\mathcal{X}\mid h_s\geq 0\},
    \label{eq:safe_set}
\end{equation}
where $h_s$ is a continuously differentiable safety barrier constructed from the current Gaussian map. Enforcing the corresponding CBF  condition renders $\mathcal{C}_s$ forward invariant, and therefore keeps the robot safe with respect to the learned scene representation~\cite{cbf_aron2019}.
In parallel, the robot should acquire observations that reduce uncertainty in regions relevant to its future motion. We encode this objective with a perception barrier $h_p$, whose value measures whether the current state and sensing direction provide sufficient information for improving the map.

The safety and perception conditions constrain the same control input and can therefore become incompatible. A robot may need to turn away from an uncertain region to remain safe, while the perception objective may favor viewing that same region to reduce uncertainty. Treating both requirements as hard constraints can therefore leave no admissible action in cluttered or partially observed scenes. Since collision avoidance is a physical requirement and information gathering is a mission objective, we enforce safety as a hard constraint and allow perception to be relaxed when there is a conflict with the safety.

\textbf{Problem Statement.}
Given the current robot state $x_k$, the attributed Gaussian map $\mathcal{G}_k$, the nominal control policy  $u_k^{\mathrm{ref}}$, and the admissible control set $\mathcal{U}$, compute the next best safe-informative action policy $u_k^\star$ by solving
\begin{equation}
\begin{aligned}
\arg \underset{u_k, \delta_p}{\text{minimize}}\quad
    &
    \frac{1}{2}\|u_k-u_k^{\mathrm{ref}}\|^{2}
    +
    \frac{1}{2}\xi_k\delta_p^2
    \\
    \mathrm{s.t.}\quad
    &
    \dot h_s
    \geq
    -\gamma_s\alpha_s\!\left(h_s\right),
    \\
    &
    \dot h_p
    \geq
    -\gamma_p\alpha_p\!\left(h_p\right)-\delta_p,
    \\
    &
    u_{\min} \preceq u \preceq u_{\max},\qquad \delta_p\geq0 ,
\end{aligned}
\label{eq:problem_statement}
\end{equation}
where $\alpha_s$ and $\alpha_p$ are extended class-$\mathcal{K}$ functions.
The safety constraint is imposed as a hard, unrelaxed constraint. The perception slack $\delta_p$ is penalized by the adaptive weight $\xi_{k}$. The objective is to design $h_s^k$, $h_p^k$, and $\xi_{k}$ such that the robot preserves collision avoidance while actively improving the Gaussian map along its task-relevant future motion.

\section{Safe-informative action CBF design}


\subsection{Safety Control Barrier Function}
\label{subsec:safety_cbf}

The safety objective is to prevent the robot from executing control actions
that reduce its clearance to uncertain geometric regions represented in the
current 3D Gaussian map. To account for safe navigation, we model the
distance between the robot position $\pi$ and the center of the $i$-th
Gaussian as a random variable. We quantify the
lower-tail collision risk using the Average Value-at-Risk (AV@R). 
Following~\cite{liu2024riskawarefisherRF}, we use the isotropic approximation
$\sigma_i^2=\lambda_{\max}(\Sigma_i)$,
accordingly, $d_i(\pi)\sim\mathcal{N}\!\left(\|\mu_i-\pi\|_2,\sigma_i^2\right).$
For \(\epsilon\in(0,1)\), its lower-tail AV@R is defined as
\begin{equation}
    \operatorname{AV@R}_{\varepsilon}\!\left(d_i(\pi)\right)
    :=
    \mathbb{E}\!\left[
        d_i(\pi)
        \,\middle|\,
        d_i(\pi)
        <
        \operatorname{VaR}_{\varepsilon}\!\left(d_i(\pi)\right)
    \right],
    \label{eq:avar_definition}
\end{equation}
where 
\begin{equation}
    \operatorname{VaR}_{\varepsilon}\!\left(d_i(\pi)\right)
    :=
    \inf
    \left\{
        \zeta\in\mathbb{R}
        \,\middle|\,
        \mathbb{P}\!\left(d_i(\pi)\leq\zeta\right)
        \geq\varepsilon
    \right\}
    \label{eq:var_definition}
\end{equation}
is the corresponding Value-at-Risk. We then define the risk-aware
clearance between the robot and Gaussian $g_i$ as
\begin{equation}
    \rho_i(\pi)
    =
    \operatorname{AV@R}_{\varepsilon}\!\left(d_i(\pi)\right)
    -
    r_{\mathrm{rob}},
    \label{eq:avar_clearance}
\end{equation}
where $r_{\mathrm{rob}}$ is an inflated robot radius~\cite{khass2025active}. A nonnegative \(\rho_i(\pi)\) indicates that the lower-tail expected distance from Gaussian \(g_i\) exceeds the inflated robot radius. However, enforcing a separate safety constraint for every Gaussian is computationally prohibitive in dense 3DGS maps.

A sufficient safety condition is that the robot maintain nonnegative
risk of collision from every Gaussian in the current map $\min_{g_i\in\mathcal{G}_k}
    \rho_i(\pi)
    \geq 0.$
However, the pointwise minimum is nonsmooth and may switch between active
Gaussian primitives as the robot moves. We therefore replace it with the log-sum-exp approximation
\begin{equation}
    h_s(\pi)
    =
    -\frac{1}{\beta_s}
    \log\!\left(
        \sum_{g_i\in\mathcal{G}_k}
        e^{\ \!\left(
            -\beta_s\rho_i(\pi)
        \right)}
    \right),
    \qquad
    \beta_s>0 .
    \label{eq:geometric_safety_lse}
\end{equation}
For a finite map $\mathcal{G}_k$ with $M_k$ Gaussian primitives, this
soft-min barrier satisfies~\cite{molnar2023composing}
\begin{equation}
    \min_{g_i\in\mathcal{G}_k}\rho_i(\pi)
    -
    \frac{\ln(M_k)}{\beta_s}
    \leq
    h_s(\pi)
    \leq
    \min_{g_i\in\mathcal{G}_k}\rho_i(\pi).
    \label{eq:logsumexp_bounds}
\end{equation}
Hence, $h_s^k(\pi)\geq0$ is a smooth sufficient condition for maintaining
nonnegative risk-adjusted clearance from all Gaussian primitives. Increasing
$\beta_s$ makes $h_s$ more closely approximate the minimum clearance,
while finite $\beta_s$ preserves continuous differentiability
~\cite{khass2026conflict,molnar2023composing}.

The corresponding safe set is $ \mathcal{C}_s
    =
    \left\{
        x\in\mathcal{X}
        \,\middle|\,
        h_s\!\left(\pi\right)\geq0
    \right\}.$

For the control-affine dynamics in \eqref{eq:dynamics}, the derivative of
the safety barrier is
\begin{equation}
    \dot h_s(\pi)
    =
    L_f h_s(\pi)
    +
    L_g h_s(\pi)u .
    \label{eq:safety_barrier_derivative}
\end{equation}
We impose the standard control barrier function condition
\begin{equation}
    L_f h_s(\pi)
    +
    L_g h_s(\pi)u
    \geq
    -\gamma_s\alpha_s\!\left(h_s(\pi)\right),
    \label{eq:safety_cbf_constraint}
\end{equation}
where $\alpha_s(\cdot)$ is an extended class-$\mathcal{K}$ function. This constraint is
included in the local control optimization as a hard constraint and is not
relaxed by the perception slack variable. Therefore, if the robot starts in
$\mathcal{C}_s$ and the safety constraint remains feasible at each control
update, the resulting closed-loop system preserves the safe set
over the corresponding control intervals by the forward invariance property
of CBFs~\cite{cbf_aron2019}.

The log-sum-exp structure also provides safety-attention weights for safety-critical control. Differentiating \eqref{eq:geometric_safety_lse} with respect to the robot position gives
\begin{equation}
\begin{aligned}
&\nabla_{\pi} h_s(\pi)
=
\sum_{g_i\in\mathcal{G}_k}
w_i(\pi)\,\nabla_{\pi}\rho_i(\pi),
\\
&w_i(\pi)
=
\frac{e^{\ \!\left(-\beta_s\rho_i(\pi)\right)}}
{\sum_{g_j\in\mathcal{G}_k}
e^{\ \!\left(-\beta_s\rho_j(\pi)\right)}},
\label{eq:hs_gradient}
\end{aligned}
\end{equation}
where \(w_i(\pi)\geq 0\) and \(\sum_i w_i(\pi)=1\). Thus, the composite safety gradient is a convex combination of risk of collision to the individual Gaussian gradients. The weights increase for Gaussians with smaller clearance, causing the barrier gradient to be dominated by the most critical nearby obstacles while remaining continuously differentiable. This provides a smooth and geometrically consistent safety direction for the CBF controller and avoids abrupt switching between active Gaussian primitives.

\subsection{Active-Perception Control Barrier Function}
\label{subsec:active_perception_cbf}

The safety barrier constrains the robot to avoid hazardous regions of the current Gaussian map. Still, safe motion alone does not ensure that the robot acquires observations useful for future planning. In partially observed environments, the robot should not spend sensing effort uniformly over the entire scene; instead, it should reduce uncertainty in regions that are relevant to its upcoming motion. We therefore define a trajectory relevant perception region from the nominal rollout induced by $u_{\mathrm{ref}}$. Let
$\mathcal{P}_k=\{\bar p_0,\bar p_1,\ldots,\bar p_H\}$
denote the predicted robot positions over a short horizon, with $\bar p_0=\pi(x_k)$. 
We define an adaptive masking radius at each waypoint by
\begin{equation}
    r_\tau(\bar p_\tau)
    =
    r_{\min}
    +
    \beta_1
    \exp\!\left(
        -\beta_2  \min_{g_i\in\mathcal{G}_k} (\rho_i(\bar p_\tau))
    \right),
    \label{eq:risk_mask_radius}
\end{equation}
where $\beta_1,\beta_2>0,$ and $r_{\min}\ge0$ is a minimum look ahead radius.
The local perception mask is
\begin{equation}
    \Omega_k
    =
    \bigcup_{\tau=0}^{H}
    \mathcal{B}(\bar p_\tau,r_\tau),
    \label{eq:perception_mask}
\end{equation}
where $\mathcal{B}(\bar p_\tau,r_\tau)$ is a ball centered at the nominal future position. The radius $r_\tau$ is enlarged in regions of higher collision risk, so perception is concentrated on the parts of the scene most likely to affect the risk of collision in future robot motion. The active perception set is then
\begin{equation}
    \mathcal{G}_{k|\Omega}
    =
    \{g_s\in\mathcal{G}_k \mid \mu_s\in\Omega_k\},
    \label{eq:masked_gaussian_set}
\end{equation}
which makes the information evaluated only over the Gaussian primitives that lie near the predicted robot trajectory.


We quantify the value of a sensing pose by the expected reduction in uncertainty of the Gaussian parameters inside the trajectory-relevant mask. Let $w_{\Omega}$ denote the subset of 3DGS parameters associated with $\mathcal{G}_{k|\Omega}$, and let $Y$ be the RGB-D observation predicted from sensing pose $(\pi,\eta)$. The information value is the conditional mutual information
\begin{equation}
    \mathcal{I}(w_{\Omega};Y\mid \pi,\eta)
    =
    \mathbb{H}(w_{\Omega})
    -
    \mathbb{H}(w_{\Omega}\mid Y,\pi,\eta),
    \label{eq:masked_mutual_information}
\end{equation}
where $\mathbb{H}(\cdot)$ denotes entropy~\cite{kirsch2022unifying}. 
The observed Fisher matrix at the
candidate view is
\begin{equation}
\small
\begin{aligned}
H''\!\left[Y\mid(\pi_i,\eta_i),\mathcal{G}_{k|\Omega}\right]
    :=
    -\nabla_{w_{\Omega}}^2
    \log p\!\left(Y\mid(\pi_i,\eta_i),w\right)
    \big|_{w_{\Omega}=w_{\Omega}^\star},
    \label{eq:observed_information}
\end{aligned}
\end{equation}
Let $H''[\mathcal{G}_{k|\Omega}]_{\mathrm{prior}}\succ0$ denote the regularized accumulated information from previous observations. With the diagonal Gauss--Newton Fisher approximation, the trajectory-relevant expected information gain is~\cite{jiang2024fisherrf}

\begin{equation}
    \mathcal I
    \left(
        (\pi,\eta)
    \right)
    \approx
    \operatorname{tr}
    \left(
        \bar H''
        \left[
            Y\mid(\pi,\eta),\mathcal{G}_{k|\Omega}
        \right]
        H''
        \left[
            \mathcal{G}_{k|\Omega}
        \right]_{\mathrm{prior}}^{-1}
    \right).
    \label{eq:trajectory_relevant_eig}
\end{equation}
Here, the prior inverse represents the remaining uncertainty in the masked Gaussian map, while the observed information measures how strongly the candidate view constrains that uncertainty. Therefore, the score is high when the robot is expected to observe uncertain Gaussian parameters near its future trajectory, rather than merely improving global reconstruction quality~\cite{khass2025active,mollaeikhass2026semsafe3dgs}.

We convert the information acquisition into a perception barrier by requiring the current sensing configuration to maintain a minimum level of trajectory-relevant information:
\begin{equation}
    h_p(\pi,\eta)
    =
    \mathcal I
    \left(
        (\pi,\eta)
    \right)
    -
    \mathcal I_c ,
    \label{eq:perception_barrier_fisher}
\end{equation}
where $\mathcal I_c>0$ is the desired information threshold. The active perception CBF condition is then
\begin{equation}
    \dot h_p(\pi,\eta) =     L_fh_p
    +
    L_gh_p\,u
    \geq
    -\gamma_p
    \alpha_p
    \left(
        h_p(\pi,\eta)
    \right)
    -
    \delta_p,
    \
    \delta_p\geq0 .
    \label{eq:active_perception_cbf_fisher}
\end{equation}
Unlike the safety CBF, the perception CBF is softened by the slack variable $\delta_p$. Informative sensing is valuable for reducing uncertainty and improving future decisions, but it must be relaxed whenever it conflicts with collision avoidance, actuation limits, or task progress.


\subsection{Unified Conflict Aware CBF-QP Policy }
\label{subsec:conflict_qp}

The safety and perception barriers define two desired properties of the robot motion. The safety barrier prevents the robot from approaching collision with risky Gaussian regions, while the perception barrier encourages actions that acquire informative observations for future planning.

The safe-informative control action is obtained from optimization problem~\eqref{eq:problem_statement}.
To make the conflict aware algorithm explicit, consider the Lagrangian of \eqref{eq:problem_statement}:
\begin{equation}
\begin{aligned}
&\mathcal{L}(u,\delta_p,\lambda_s,\lambda_i,\nu_{p}, \mu_i^+, \mu_i^-)
=
\frac{1}{2}\|u-u_{\mathrm{ref}}\|^2
    +\frac{1}{2}\xi_k\delta_p^2
    \\
&
+\lambda_s\!\left(
- L_f h_s - L_g h_su - \gamma_s\alpha_s(h_s)
\right)
    \\
&
+\lambda_p\!\left(
- L_f h_p - L_g h_p u - \gamma_{p} \alpha_p (h_p) - \delta_p
\right)
    \\
&
    +(\mu^+)^\top(u-u_{\max})
    +(\mu^-)^\top(u_{\min}-u)
    \\
&
    +\nu(-\delta_p) ,
\end{aligned}
\label{eq:lagrangian}
\end{equation}
where $\lambda_s,\lambda_p,\nu\geq0$ and $\mu^+,\mu^-\succeq0$ are the multipliers associated with safety, perception, slack nonnegativity, and input bounds.
For simplicity of our notations, we denote $A_{\star} = L_g h_{\star}$ and $b_{\star} = -L_f h_{\star} - \gamma_{\star} \alpha_{\star}(h_{\star})$ for each $\star \in \{s, p\}$. 
The stationarity conditions are
\begin{equation}
\small
\begin{aligned}
    u^\star= u_{\mathrm{ref}}
    +A_s^\top\lambda_s
    +A_p^\top\lambda_p
    -\mu^++\mu^-,\ \xi_k\delta_p^\star= \lambda_p + \nu.
    \label{eq:kkt_u}
\end{aligned}
\end{equation}
Together with primal feasibility, the complementary slackness conditions are
\begin{equation}
\begin{aligned}
    \lambda_s(A_su^\star-b_s)=0,
    \
    \lambda_p(A_pu^\star+\delta_p^\star-b_p)=0, \
    \nu\delta_p^\star=0 .
\end{aligned}
\label{eq:kkt_complementarity}
\end{equation}
Since the safety constraint has no slack, its multiplier $\lambda_s$ is not capped by any relaxation variable and can dominate the control correction whenever safety becomes active. By contrast, if the perception constraint is violated, then $\delta^\star>0$, $\nu=0$, and \eqref{eq:kkt_u} gives $    \lambda_p = \xi_k\delta_p^\star.$

Therefore, the perception multiplier is proportional to both the amount of perception violation and the slack weight. The slack weight is consequently not a numerical tuning artifact; it directly controls the marginal cost of sacrificing perception in the optimality conditions.

A fixed slack weight cannot express the changing value of information during navigation. Early relaxation of perception is often harmless, but if the robot repeatedly sacrifices perception it accumulates unresolved uncertainty in the very regions that shape its future motion, weakening the reliability of the safety margins it
depends on. The penalty on the perception slack should therefore be small when information is cheap to forgo and large exactly when neglecting perception would leave the robot poorly informed near risk. Since the safety constraint is never relaxed, this weight does not affect forward invariance; it only arbitrates how aggressively perception is preserved when it competes with the nominal command.

We drive the weight with two complementary cues. The first is a history of past relaxation, captured by a leaky integrator of the realized slack,
\begin{equation}
  \bar{\delta}_k = \lambda\,\bar{\delta}_{k-1} + \delta_{k-1}^{\star},
  \qquad \lambda \in (0,1),
\end{equation}
which grows when perception has been repeatedly sacrificed and decays otherwise. The second is the robot's proximity to the map frontier. As the trajectory approaches unexplored space,
$d_k = \min_{\tau} d_{\mathrm{front}}(\bar{p}_\tau)$ shrinks, and the value of acquiring information before entering that region rises. Both cues raise a single nonnegative urgency score
\begin{equation}
  z_k = \mu\,\bar{\delta}_k + a\,e^{-d_k/d_0},
\end{equation} 
so that recent neglect and frontier proximity reinforce one another.
The slack weight is then obtained by a bounded, monotone map
\begin{equation}
  \xi_k = \xi_{\min} + (\xi_{\max}-\xi_{\min})\bigl(1 - e^{-z_k}\bigr),
  \ 0 < \xi_{\min} < \xi_{\max},
\end{equation}
which keeps $\xi_k \in [\xi_{\min},\xi_{\max}]$ by construction and increases with both cues. When no perception has recently been lost, and the frontier is distant, $\xi_k \to \xi_{\min}$ and perception is
readily relaxed in favor of the nominal task. When the robot has been neglecting perception and is approaching the unknown, $\xi_k \to \xi_{\max}$, and perception is strongly protected, the robot reduces uncertainty before committing to risky future motion. 


\section{Experiments}
\label{sec:experiments}

\begin{table}[t]
\centering
\vspace{1.8mm}
\scriptsize
\setlength{\tabcolsep}{2.1pt}
\renewcommand{\arraystretch}{0.80}
\resizebox{\columnwidth}{!}{%
\begin{tabular}{@{}llcccc@{}}
\toprule
\textbf{Scene}
& \textbf{Method}
& \shortstack{\textbf{Safety}\\\textbf{[\%] $\uparrow$}}
& \shortstack{\textbf{Success}\\\textbf{[\%] $\uparrow$}}
& \shortstack{\textbf{Time}\\\textbf{[ms] $\downarrow$}}
& \shortstack{\textbf{Min. dist.}\\\textbf{}} \\
\midrule

\multirow{4}{*}{\shortstack[l]{Stonehenge\\{\footnotesize $1.16\times10^5$ Gs}}}
& PolyMerge
& 95 & 96 & 1.17 & 0.55 \\
& SAFER-Splat
& 100 & 97 & 33.33 & 0.47 \\
& \textsc{Splat-CBF} (Ours), $\varepsilon=.30$
& 100 & 99 & 1.01 & 0.73 \\
& \textsc{Splat-CBF} (Ours), $\varepsilon=.10$
& 100 & 77 & 1.00 & 0.98 \\
\midrule

\multirow{4}{*}{\shortstack[l]{Statues\\{\footnotesize $2.02\times10^5$ Gs}}}
& PolyMerge
& 99 & 100 & 1.16 & 2.80 \\
& SAFER-Splat
& 99 & 98 & 47.16 & 1.73 \\
& \textsc{Splat-CBF} (Ours), $\varepsilon=.30$
& 99 & 96 & 1.00 & 1.86 \\
& \textsc{Splat-CBF} (Ours), $\varepsilon=.10$
& 99 & 96 & 1.00 & 2.31 \\
\midrule

\multirow{4}{*}{\shortstack[l]{Flight\\{\footnotesize $2.82\times10^5$ Gs}}}
& PolyMerge
& 94 & 94 & 1.16 & 1.38 \\
& SAFER-Splat
& 99 & 89 & 67.00 & 0.85 \\
& \textsc{Splat-CBF} (Ours), $\varepsilon=.30$
& 99 & 85 & 1.00 & 1.20 \\
& \textsc{Splat-CBF} (Ours), $\varepsilon=.10$
& 99 & 64 & 1.01 & 1.71 \\
\midrule

\multirow{4}{*}{\shortstack[l]{Old Union\\{\footnotesize $5.26\times10^5$ Gs}}}
& PolyMerge
& 99 & 92 & 1.16 & 1.39 \\
& SAFER-Splat
& 100 & 70 & 107.79 & 0.37 \\
& \textsc{Splat-CBF} (Ours), $\varepsilon=.30$
& 100 & 93 & 0.96 & 0.45 \\
& \textsc{Splat-CBF} (Ours), $\varepsilon=.10$
& 100 & 97 & 0.98 & 0.70 \\
\bottomrule
\end{tabular}%
}
\caption{Safety-navigation comparison across four 3DGS scenes. Time denotes the mean online control computation per step.}
\vspace{-6.5mm}
\label{tab:safety_comparison}
\end{table}
We evaluate the proposed framework through simulation and hardware experiments. First, we compare the risk-aware safety CBF in Sec.~\ref{subsec:safety_cbf} against SAFER-Splat~\cite{chen2025control} and PolyMerge~\cite{hong2026polymerge} in terms of safety, task success, and computation. We then evaluate the Splat-CBF method against CAAP~\cite{khass2026conflict} and passive perception mapping, followed by ablations of the adaptive relaxation mechanism. Finally, we validate the framework on an Ackermann-steered robot and an Isaac Sim Kinova manipulator simulation to evaluate closed-loop safety--perception conflicts and map improvement. In our experiments, \(\mathcal G_k\) is held fixed during each control interval. 
Experiments A--D were conducted on an AMD Ryzen~7 7700 CPU with an NVIDIA
GeForce RTX~3090 GPU and Experiment~E used an Intel Core i9-13900K CPU with an NVIDIA RTX~A2000 GPU.

\subsection{Safety Barrier Evaluation}
\label{sec:exp_safety}


\subsubsection{Experimental Setup}
\label{sec:exp_safety_setup}
We evaluate the proposed safety barrier on four 3DGS scenes against
SAFER-Splat~\cite{chen2025control}, which imposes per-Gaussian CBF
constraints, and PolyMerge~\cite{hong2026polymerge}, which first converts
the map into compact polytopes. All methods use the same queries, planning
settings, spherical robot, double-integrator dynamics, control limits, and
collision checker. A waypoint-tracking PD controller provides the nominal
input, which is modified by each safety filter. We conduct $100$ start--goal
trials per scene and evaluate our method with
$\varepsilon\in\{0.30,0.10\}$ and $\beta_s=300$.
Safety is the percentage of trials maintaining nonnegative
clearance throughout execution; success is the percentage terminating within
$10\%$ of the initial start--goal distance from the goal. Time denotes the
mean online control computation per step.



\subsubsection{Results}
\label{sec:exp_safety_results}
\begin{figure}[t]
    \centering
    \vspace{1.8mm}
    \includegraphics[
width=0.7\columnwidth,height=0.14\textheight,
        trim={0.25cm 0.30cm 0.90cm 0.15cm},
        clip
    ]{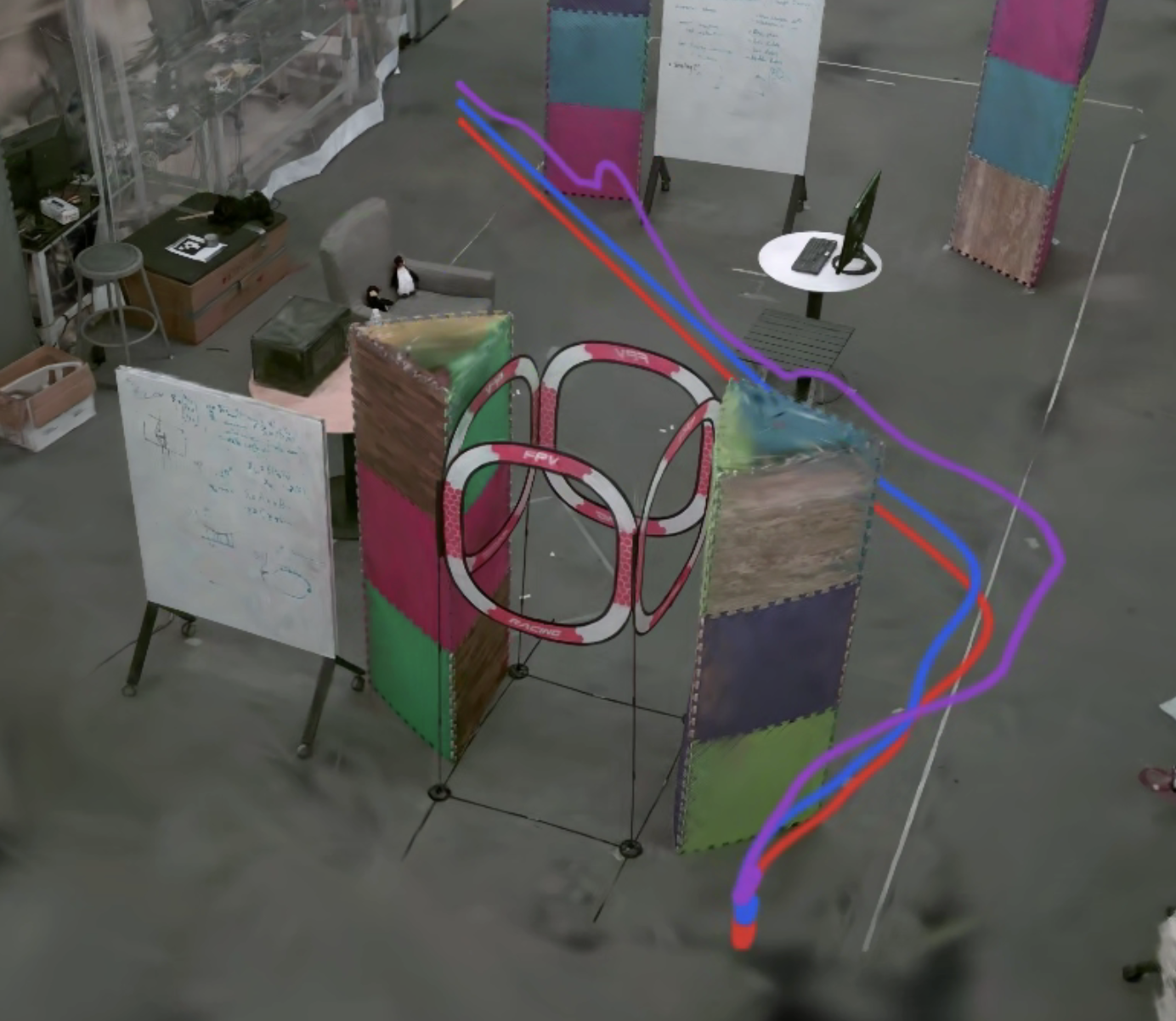}
\caption{Qualitative comparison of safety-navigation trajectories in a 3DGS environment: Splat-CBF (ours, \textcolor{red}{\rule{2mm}{2mm}}), SAFER-Splat (\textcolor{blue}{\rule{2mm}{2mm}}), and PolyMerge (\textcolor{purple!99!black}{\rule{2mm}{2mm}}).}
    \label{fig:trajcomp}
    \vspace{-3.5mm}
\end{figure}

Table~\ref{tab:safety_comparison} shows that both Splat-CBF settings achieve
$99.5\%$ safety, matching SAFER-Splat and improving upon PolyMerge
($96.8\%$). With $\varepsilon=.30$, Splat-CBF provides the best overall
balance, approaching PolyMerge's success rate ($93.3\%$ versus $95.5\%$)
while offering higher safety and outperforming SAFER-Splat in success
($88.5\%$). The more conservative $\varepsilon=.10$ setting increases the
minimum clearance in every scene, but reduces success to $83.5\%$.
Moreover, at the selected polytope resolution, PolyMerge requires
$0.55$--$1.34$\,min of offline map compression before deployment; this
preprocessing cost is excluded from its reported online runtime.
Splat-CBF also reduces the mean online computation from $63.82$\,ms for
SAFER-Splat to $0.99$\,ms, yielding an approximately $64\times$ speedup.
This efficiency follows from replacing numerous obstacle-wise constraints
with a single differentiable composite barrier. Fig.~\ref{fig:trajcomp}
qualitatively confirms that this compact formulation produces collision-free
trajectories through cluttered 3DGS environments.

\vspace{-1.5mm}
\subsection{Trajectory-Aware Active Perception}
\label{sec:exp_trajectory_aware}

\begin{figure*}[t]
    \centering
    \vspace{0.8mm}
    \includegraphics[width=\textwidth,trim={3.4cm 6.7cm 3.4cm 4.3cm},
    clip]{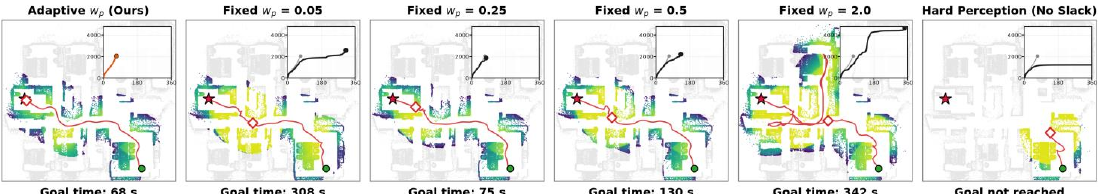}
\caption{Ablation of the perception-slack weighting using adaptive, fixed,
and hard formulations. The \textcolor{red}{$\Diamond$} denotes the robot pose at $t=68$\,s, when
the adaptive method reaches the goal. Insets show cumulative uncertainty
reduction over time, with the adaptive result in gray as reference.}
    \label{fig:slack_weight_comparison}
\end{figure*}

\begin{figure}[t]
    \centering
    \includegraphics[
width=0.6\textwidth,
        trim={5.6cm 4.6cm 2.7cm 3.7cm},
        clip
    ]{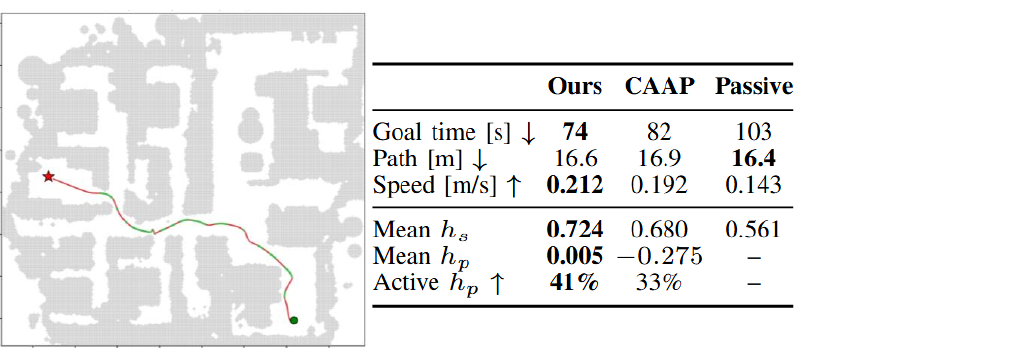}
\vspace{-20pt}
\caption{The Splat-CBF trajectory
(left) Green (\textcolor{green}{\rule{2mm}{2mm}}) denotes compatible safety and
perception objectives ($\delta_p=0$), while red
(\textcolor{red}{\rule{2mm}{2mm}}) indicates conflicts resolved by relaxing
perception ($\delta_p>0$) while preserving hard safety.
Compared with CAAP and
passive mapping, Splat-CBF reaches the goal fastest while maintaining
positive mean safety and perception barriers.}
\label{fig:exp2explain}
\end{figure}

\subsubsection{Experimental Setup}
\label{sec:exp_trajectory_aware_setup}
We evaluate the proposed trajectory-aware perception CBF against CAAP~\cite{khass2026conflict} and a safety-only passive-mapping baseline in InteriorGS~\cite{InteriorGS2025}. CAAP uses fixed-weight spatial and angular CBFs to align motion with local EIG gradients. The passive baseline retains the same hard safety CBF but omits active perception, updating the map only from observations encountered during navigation. All methods use identical start--goal states, unicycle dynamics, control limits, nominal $A^\star$ planning, sensing, and online mapping.

At each control step, one RGB-D observation updates the 3DGS map and its uncertainty estimates. The nominal path is then replanned, both barriers are recomputed from the updated map, and the CBF-QP generates the next control input. Thus, the map is updated once per control step.

\subsubsection{Results}
\label{sec:exp_trajectory_aware_results}

Fig.~\ref{fig:exp2explain} summarizes the navigation and perception results.
Splat-CBF reaches the goal in $74$\,s, compared with $82$\,s for CAAP and
$103$\,s for passive mapping. The three methods traverse comparable distances
($16.6$, $16.9$, and $16.4$\,m, respectively); nevertheless, Splat-CBF
achieves the highest mean speed, $0.212\,\mathrm{m/s}$. In particular,
although passive mapping follows a marginally shorter path, its limited
information acquisition leaves future regions insufficiently resolved. As
the risk-aware clearance decreases, the hard safety CBF must therefore
attenuate the nominal velocity more frequently, resulting in slower progress.

Splat-CBF avoids this behavior by introducing short, informative deviations
that improve the map along the predicted trajectory. These observations
reduce uncertainty before the robot enters safety-critical regions, allowing
the controller to sustain faster motion without sacrificing clearance. This
is reflected by the largest mean safety barrier,
$\operatorname{mean}(h_s)=0.724$, and a positive mean perception barrier,
$\operatorname{mean}(h_p)=0.005$. CAAP improves upon passive mapping,
but it does not directly regulate the
information available along the future trajectory which yields
$\operatorname{mean}(h_p)=-0.275$ due to need of exploration and lack of adaptive slack penalty weight.

\begin{figure}[t]
    \centering
    \includegraphics[
width=0.90\columnwidth,height=0.16\textheight,
        trim={0.55cm 0.30cm 0.90cm 0.25cm},
        clip
    ]{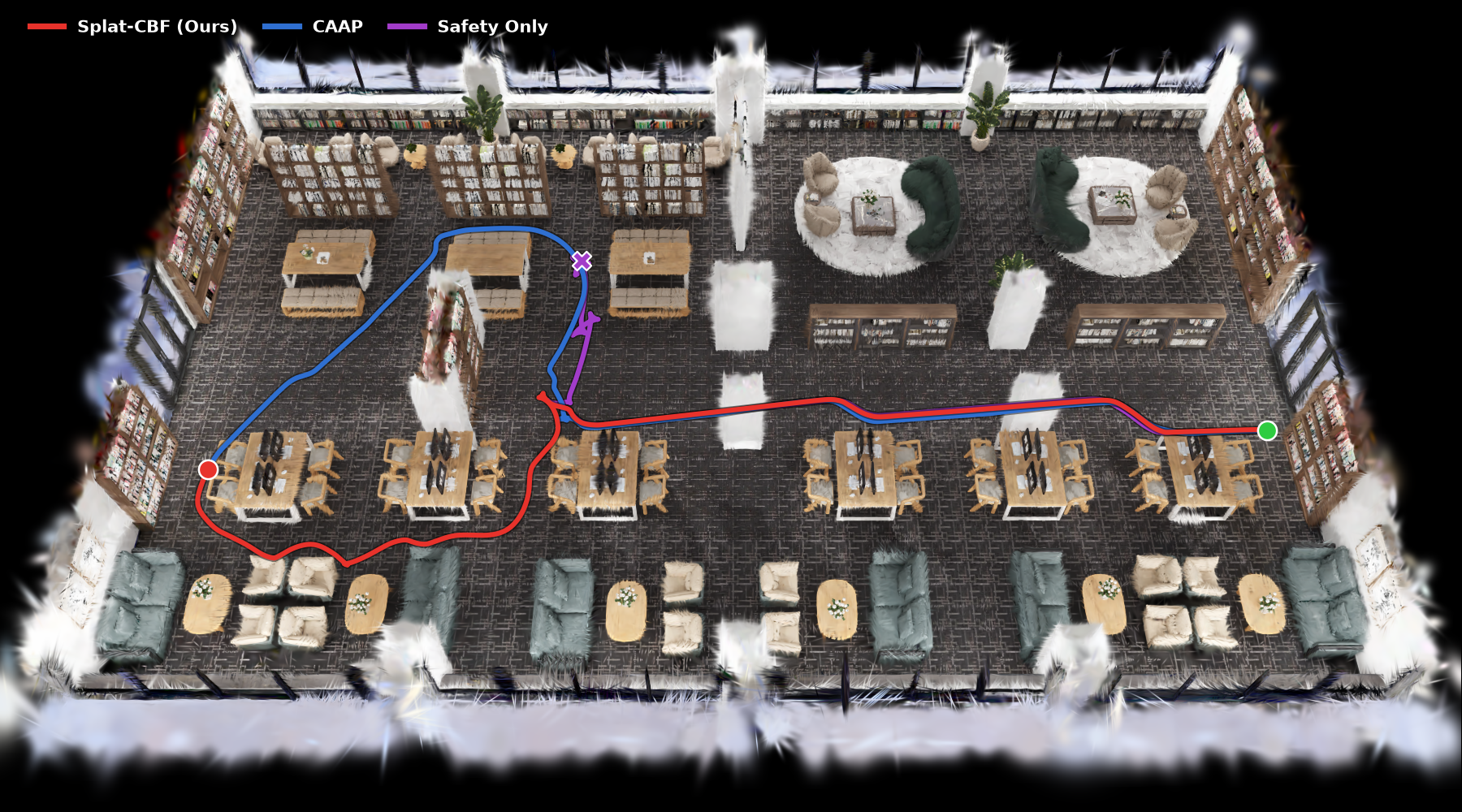}
\caption{Qualitative trajectory comparison in the same 3DGS environment. From the common start (green), Splat-CBF (ours, \textcolor{red}{\rule{2mm}{2mm}}) and CAAP (\textcolor{blue}{\rule{2mm}{2mm}}) reach the goal (red), whereas Safety Only (\textcolor{purple!75}{\rule{2mm}{2mm}}) terminates before reaching it.}
    \label{fig:trajecortyCAAD}
    \vspace{-3.5mm}
\end{figure}

Fig.~\ref{fig:trajecortyCAAD} provide
complementary views of this behavior. It compares the complete trajectories and shows that passive mapping can lose progress when its
nominal command repeatedly directs the robot toward poorly observed,
safety-limited regions. 
Fig.~\ref{fig:exp2explain} resolves the proposed trajectory into intervals of safety--perception agreement and conflict. When the
objectives are compatible, perception is enforced without relaxation
($\delta_p=0$, green); during conflicts, the hard safety constraint is
retained while perception is relaxed through $\delta_p>0$ (red). Perception
remains unrelaxed for $41\%$ of the Splat-CBF trajectory, compared with
$33\%$ for CAAP, demonstrating more persistent information acquisition while preserving hard safety.

\subsection{Ablation Study}
\label{sec:exp_slack_ablation}

We compare the proposed adaptive slack weight with fixed penalties
$w_p\in\{0.05,0.25,0.5,2.0\}$ and a hard perception constraint without
slack. All configurations use the same initial map, start--goal pair,
planner, safety CBF, and perception objective; only the perception-relaxation
mechanism changes. Fig.~\ref{fig:slack_weight_comparison} reports executions configuration and compares the resulting trajectories,
goal times, and cumulative uncertainty reduction.

Without relaxation, enforcing safety and perception simultaneously can render the QP
infeasible during a conflict, and the robot fails to reach the goal.
Introducing perception slack removes this source of infeasibility, but a
fixed penalty cannot respond to changes in clearance and map uncertainty.
With $w_p=0.05$, perception is inexpensive to relax; trajectory-relevant
uncertainty is therefore reduced too slowly, causing repeated intervention
by the hard safety filter and a goal time of $308$\,s. Conversely, larger
weights retain the perception objective too aggressively, producing
information-seeking detours and increasing the goal time to $130$\,s for
$w_p=0.5$ and $342$\,s for $w_p=2.0$. The intermediate setting
$w_p=0.25$ performs best among the fixed penalties, reaching the goal in
$75$\,s, but remains manually tuned to this operating condition.

The proposed method instead adapts $w_p$ using the current safety margin, map
uncertainty, proximity to poorly observed regions, and recent slack history.
Persistent relaxation increases the penalty and encourages recovery of
informative sensing once sufficient clearance becomes available. During
conflicts, the hard safety constraint remains dominant, while the slack
absorbs the perception violation.
. This state-dependent arbitration reaches the goal fastest,
in $68$\,s, while retaining meaningful uncertainty reduction. Moreover, the
larger final uncertainty reduction of some fixed-weight runs partly reflects
their substantially longer duration, whereas the adaptive method acquires
useful information more efficiently while preserving task progress.

\subsection{Real-Robot Validation with Ackermann Dynamics}
\label{sec:exp_ackermann}

\subsubsection{Experimental Setup}
\label{sec:exp_ackermann_setup}

We validate the proposed framework on a physical Ackermann-steered mobile robot navigating an indoor environment represented by an online 3DGS map. Unlike the unicycle simulations, the vehicle is subject to nonholonomic steering, bounded acceleration, and limited braking authority. We therefore use a higher-order safety CBF consistent with the vehicle dynamics while retaining the soft trajectory-aware perception CBF. A nominal trajectory provides $u_{\mathrm{ref}}$, and the unified CBF-QP modifies this command online to satisfy the steering, actuation, safety, and perception constraints.

\subsubsection{Results}
\begin{figure}[t]
    \centering
    \vspace{1.5mm}
    \includegraphics[
        width=\columnwidth,
        trim={0.15cm 0.25cm 0.15cm 0.05cm},
        clip
    ]{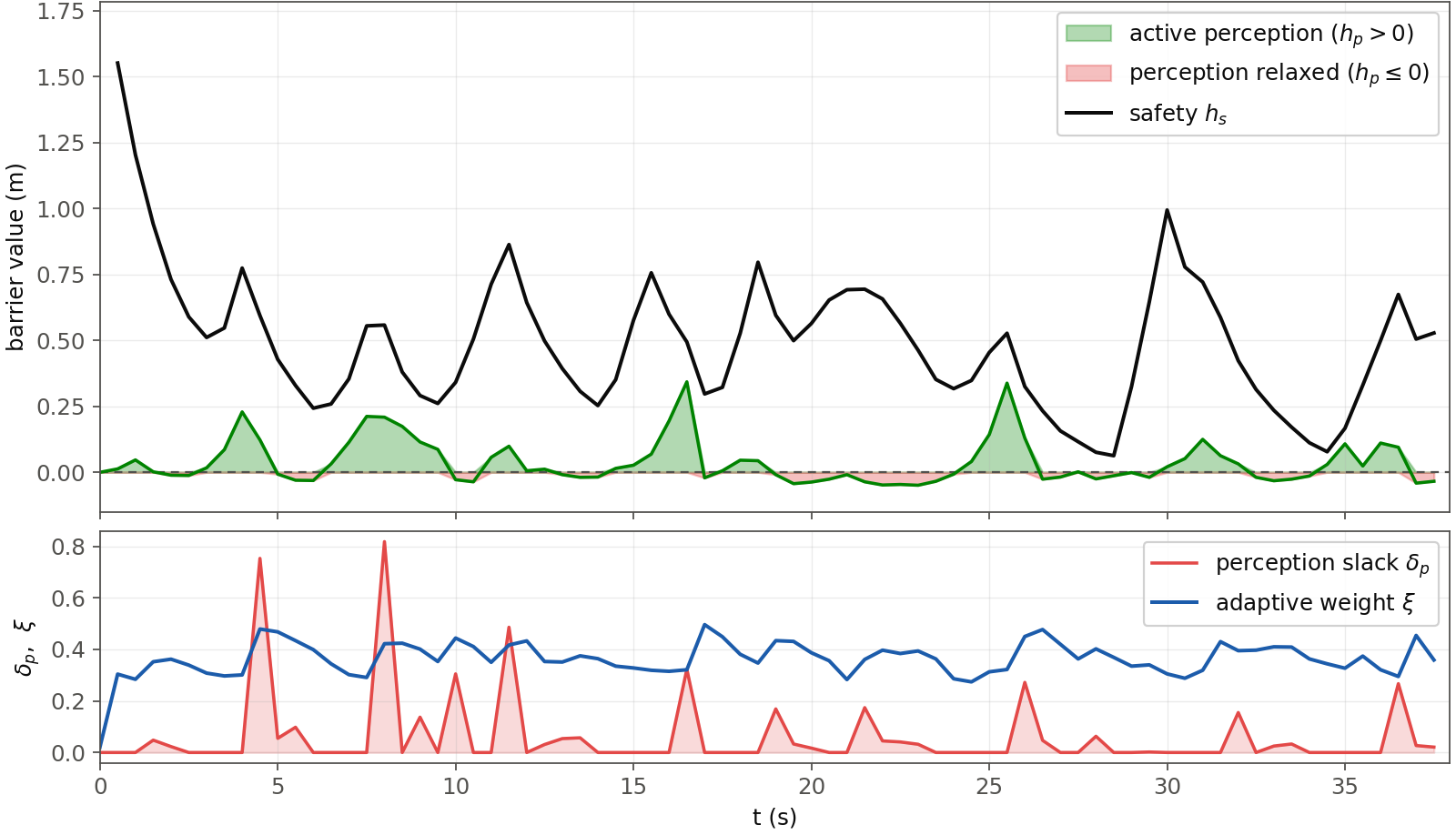}
    \vspace{-6.5mm}
\caption{Closed-loop safety--perception arbitration on the Ackermann robot. The safety barrier $h_s$ remains positive, while the perception constraint is selectively relaxed through $\delta_p$ during conflicts; the adaptive weight $\xi$ regulates the cost of this relaxation online.}
    \label{fig:ackermann_hs}
    \vspace{-1.5mm}
\end{figure}

\begin{figure}[t]
    \centering
    \includegraphics[
        width=\columnwidth,height=0.14\textheight,
        trim={0.25cm 0.60cm 0.90cm 2.6cm},
        clip
    ]{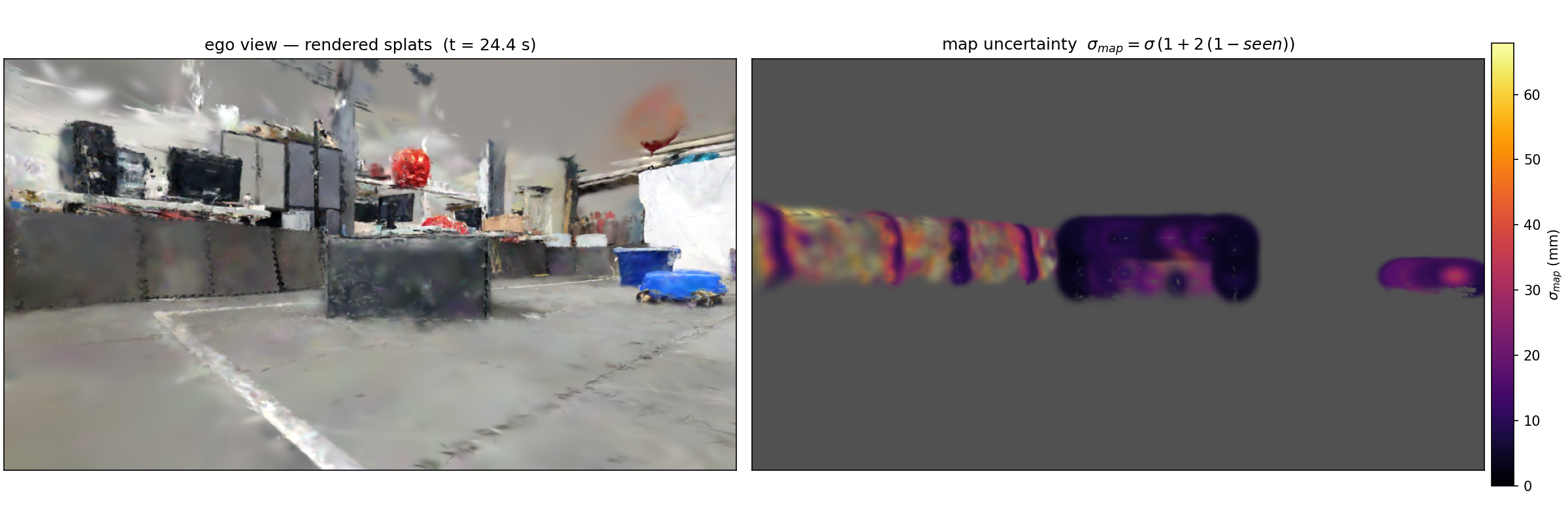}
    \vspace{-6.5mm}
\caption{Online 3DGS reconstruction during the Ackermann experiment (left) and the corresponding map uncertainty (right), highlighting poorly observed regions relevant to safe-active navigation.}
    \label{fig:ego}
    \vspace{-3.5mm}
\end{figure}

\begin{table}[t]
    \small
    \vspace{1.5mm}
    \setlength{\tabcolsep}{5pt}
    \renewcommand{\arraystretch}{1.10}

    \begin{tabular}{@{}p{0.69\columnwidth}r@{}}
        \toprule

        \multicolumn{2}{c}{\textbf{Trajectory and Safety}} \\
        \cmidrule(lr){1-2}
        \textbf{Metric} & \textbf{Value} \\
        \midrule

        Reference / executed path length [m]
            & $24.8\,/\,28.2$ \\

        Minimum / mean safety margin $h_s$ [m]
            & ${0.046}\,/\,0.492$ \\

        Cumulative online computation [s]
            & $30.1$ \\

        \midrule

        \multicolumn{2}{c}{\textbf{Perception and Mapping}} \\
        \cmidrule(lr){1-2}

        Perception-conflict duration 
            & $24.2\%$ \\

        Adaptive weight $\xi$ (min./mean/max.)
            & $0.02\,/\,0.36\,/\,0.49$ \\

        Map coverage 
            & $91\%$ \\

        \bottomrule
    \end{tabular}
        \centering
    \caption{Closed-loop Ackermann results: safety, perception-conflict, adaptive-weight, and mapping performance.}
    \vspace{-3.5mm}
    \label{tab:ackermann_results}
\end{table}

\begin{figure}[t]
    \centering
    \includegraphics[
        width=\columnwidth,
        trim={0.25cm 0.30cm 0.20cm 0.15cm},
        clip
    ]{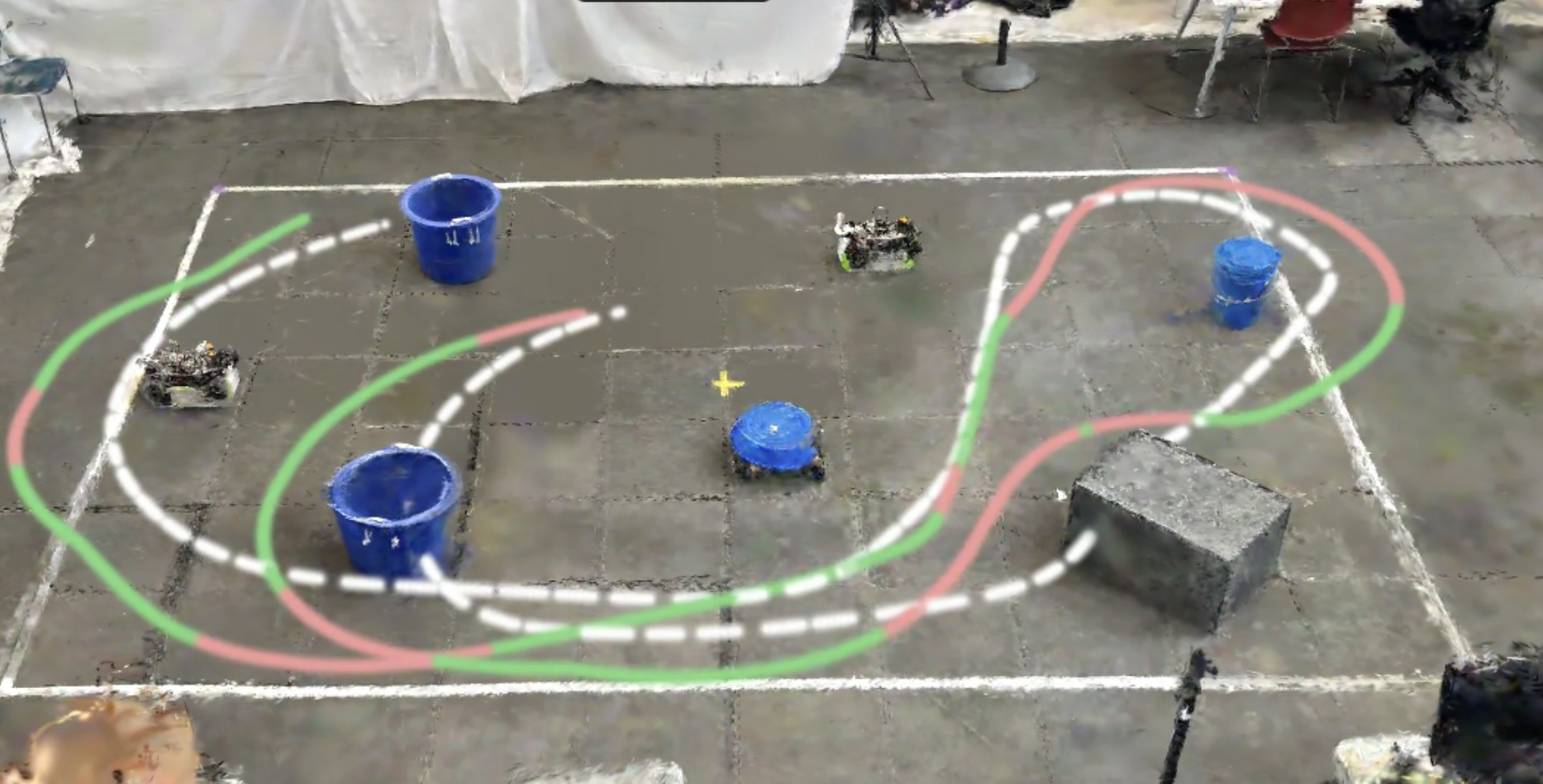}
    \vspace{-6.5mm}
\caption{Real-robot Ackermann experiment illustrating safety--perception conflict resolution. The dashed white curve denotes the nominal trajectory; Green segments (\textcolor{green}{\rule{2mm}{2mm}}) indicate compatible safety and perception objectives, while red segments (\textcolor{red}{\rule{2mm}{2mm}}) denote perception relaxation near obstacles, where the hard safety constraint takes priority.}
    \label{fig:ackermann_real_robot}
    \vspace{-4mm}
\end{figure}

\label{sec:exp_ackermann_results}
Fig.~\ref{fig:ackermann_real_robot} compares the nominal and executed
trajectories. The controller deliberately departs from the reference near
safety-critical and informative regions and returns once the constraints
become compatible. Consequently, the executed trajectory is $28.2$\,m,
compared with the nominal length of $24.8$\,m, corresponding to a $13.7\%$
overhead. This deviation reflects purposeful safety--perception intervention
rather than tracking error.

Fig.~\ref{fig:ackermann_hs} shows that $h_s$ remains positive throughout
the traversal, with minimum and mean values of $0.046$ and $0.492$\,m,
respectively. Thus, the controller preserves the safe set during the closest
obstacle encounters despite the vehicle's steering and braking limitations.
Perception conflicts, identified by $\delta_p>0$, occur during $24.2\%$ of
the run. In these intervals, only the perception constraint is relaxed,
whereas the safety CBF remains hard; once compatibility is restored,
$\delta_p$ returns toward zero and informative motion resumes.

The adaptive penalty varies from $0.02$ to $0.49$, with a mean of $0.36$,
demonstrating that the controller continuously adjusts the cost of perception
relaxation rather than imposing a fixed compromise. The experiment incurs
$30.1$\,s of cumulative online computation over the complete traversal and
achieves $91\%$ final map coverage, as reported in
Table~\ref{tab:ackermann_results}. Together, these results verify that the
proposed safety--perception arbitration operates successfully on a physical
Ackermann platform while preserving safety and recovering informative
behavior whenever sufficient control authority becomes available.

\subsection{Kinova Manipulator Evaluation}
\label{sec:kinova}

\subsubsection{Experimental Setup}
We evaluate the Splat-CBF on a simulated Kinova Gen3 seven-DoF manipulator equipped with a wrist-mounted RGB-D camera. The robot operates in Isaac Sim~\cite{NVIDIA_Isaac_Sim} model of a shelf, which contains multiple objects and partially observed regions, while Immediate3DGS~\cite{2026immediate3DGS} incrementally updates the Gaussian map from posed observations. The Kinova arm must reach a task pose near a target object while safely reducing map uncertainty for object identification or grasp preparation.
At each planning iteration, Random, COVER~\cite{chen2026coverage}, and FisherRF~\cite{jiang2024fisherrf} select three next-best views, with two transit captures and one endpoint capture per view, yielding up to nine RGB-D observations for updating the 3DGS map. A nominal operational-space controller generates the corresponding joint-velocity command. We compare two execution policies for each selector. Safety Only filters the nominal command using an environment-collision barrier constructed directly from the online 3DGS map, together with self-collision, joint-limit, and singularity barriers formulated within the operational-space CBF framework~\cite{morton2025oscbf}, whereas Splat-CBF additionally incorporates the soft perception barrier to exploit the manipulator redundancy for informative camera motion. Both methods use the same goal configurations, observation budget, motion limits, and mapping protocol. Reconstruction quality is evaluated using PSNR, SSIM, and depth RMSE.

\subsubsection{Results}
\begin{figure}[t]
    \centering
    \includegraphics[
        width=0.7\textwidth,height=0.18\textheight,
        trim={9.1cm 8.9cm 2.7cm 4.2cm},
        clip
    ]{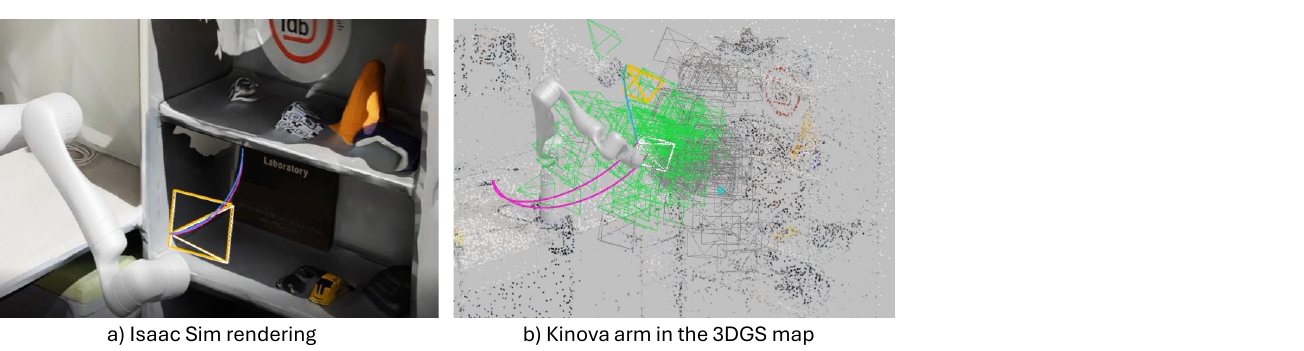}
\vspace{-20pt}
\caption{Kinova manipulator setup for safe active perception. (a) Simulated Kinova Gen3 operating in the laboratory shelf environment with the corresponding online Immediate3DGS rendering and selected camera views. (b) Kinova arm registered in the reconstructed 3D Gaussian map, showing the planned camera motion and next-best-view poses.}
\vspace{-4pt}
\label{fig:armrobot}
\end{figure}

Fig.~\ref{fig:armrobot} shows the simulated Kinova Gen3 and the reconstructed 3DGS map, together with the planned camera
motion and selected next-best-view poses.

Table~\ref{tab:armkinova} summarizes reconstruction, safety, and QP performance. Splat-CBF implementation improves reconstruction metrics across the view selectors. COVER with Splat-CBF implementation achieving the best overall result. These gains do not compromise safety; all trials remain collision-free, $h_{\mathrm{env}}^{\min}$ remains strictly positive, and Splat-CBF increases the observed minimum environment margin for every selector.
Including the active perception constraint increases the QP solve time only marginally. Table~\ref{tab:armkinova} reports the optimization time after constraint construction. 
The complete filter, including construction of the trajectory-masked Gaussian
set and finite-difference evaluation of the perception gradient, requires
approximately $0.5$--$1.5$\,s per update in the current implementation, with
runtime dominated by repeated Fisher-information evaluations.

\begin{table}[t]
\centering
\footnotesize
\setlength{\tabcolsep}{2.8pt}
\renewcommand{\arraystretch}{1.05}
\resizebox{\columnwidth}{!}{%
\begin{tabular}{@{}llccccc@{}}
\toprule
\textbf{Selector}
& \textbf{Controller}
& \makecell{\textbf{PSNR}\\$\uparrow$}
& \makecell{\textbf{SSIM}\\$\uparrow$}
& \makecell{\textbf{Depth RMSE}\\$[\mathrm{m}]\downarrow$}
& \makecell{\textbf{Minimum}\\$\mathbf{h_{\mathrm{env}}}\uparrow$}
& \makecell{\textbf{Median QP}\\
\textbf{Solve Time [ms]}$\downarrow$} \\
\midrule

\multirow{2}{*}{Random}
& Safety Only
& 15.78 & 0.649 & 0.413 & 0.126 & 0.94 \\
& Splat-CBF
& 16.85 & 0.682 & 0.381 & 0.136 & 1.13 \\
\cmidrule(lr){1-7}

\multirow{2}{*}{COVER}
& Safety Only
& 17.08 & 0.712
& 0.376 & 0.089 & 0.91 \\
& Splat-CBF
& 17.96 & 0.737
& 0.355 & 0.138 & 1.15 \\
\cmidrule(lr){1-7}

\multirow{2}{*}{FisherRF}
& Safety Only
& 16.69 & 0.667
& 0.401 & 0.077 & 0.96 \\
& Splat-CBF
& 16.87 & 0.681
& 0.393 & 0.132 & 1.18 \\

\bottomrule
\end{tabular}%
}
\caption{Reconstruction, safety, and computational performance
for different view selectors and controller variants.
}
\vspace{-13pt}
\vspace{-3.5mm}
\label{tab:armkinova}
\end{table}


\section{Conclusion}
We presented Splat-CBF, a framework for safe and informative navigation in online 3D
Gaussian maps. The method combines a risk-aware safety barrier and a trajectory-aware
perception barrier in a single CBF-QP. When the two conflict, an adaptive slack weight
relaxes perception and preserves safety. We evaluated the framework in indoor simulation, Isaac simulated Kinova manipulator and in
hardware experiments on an Ackermann-drive robot. Splat-CBF
reached its goals faster, gathered more information, and used less computation than
safety-only and full-map perception baselines, and it never collided. Future work will
accelerate the computation of the perception gradient and extend the framework to
multi-agent settings and to learning-based manipulation.

\bibliographystyle{IEEEtran}
\bibliography{references}

\end{document}